\documentclass[letterpaper, 10 pt, conference]{ieeeconf}  

\IEEEoverridecommandlockouts                              

\usepackage{amsmath} 
\usepackage{caption}
\usepackage{subcaption}
\usepackage{graphicx}
\usepackage{algorithm}
\usepackage{algpseudocode}
\usepackage[utf8]{inputenc}

\providecommand{\keyword}[1]
{
  \textbf{\textit{Keywords---}} #1
}

\title{\LARGE \bf
Enhanced $A^{*}$ Algorithm for Mobile Robot Path Planning with Non-Holonomic Constraints 
}

\author{ Suraj Kumar$^{1}$, Sudheendra R.$^{2}$, Aditya R$^{1}$,Bharat Kumar GVP$^{1}$, Ravi Kumar L$^{1}$
\thanks{$^{1}$ Controls and Digital Area, U R Rao Satellite Centre, Indian Space Research Organization (ISRO), Bangalore, India (e-mail: surajk@ursc.gov.in)%
}
\thanks{$^{2}$ Robert Bosch Centre for Cyber Physical Systems, Indian Institute of Science,
        Bengaluru, India}
}

\begin{document}

\maketitle
\thispagestyle{empty}
\pagestyle{empty}


\begin{abstract}
In this paper, a novel method for path planning of mobile robots is proposed, taking into account the non-holonomic turn radius constraints and finite dimensions of the robot. The approach involves rasterizing the environment to generate a 2D map and utilizes an enhanced version of the $A^{*}$ algorithm that incorporates non-holonomic constraints while ensuring collision avoidance. Two new instantiations of the $A^{*}$ algorithm are introduced and tested across various scenarios and environments, with results demonstrating the effectiveness of the proposed method.
\end{abstract}
\keyword{ Path planning, Non-holonomic constraints}

\section{INTRODUCTION}
\label{intro}
Path planning for mobile robots is a critical aspect of autonomous navigation, particularly when considering real-world constraints such as non-holonomic constraints and finite vehicle dimensions. The path planning module typically consists of multiple layers addressing various aspects of robot motion. The primary layer employs $A^{*}$ for path planning without considering physical constraints, while the secondary layer adjusts paths to accommodate these constraints. However, this approach often leads to dynamically inconsistent paths, such as when vehicles navigate narrow corridors, necessitating human intervention for correction and compromising autonomy.

Broadly, two major approaches for path planning are reported in the literature: grid-based and sampling-based methods. Grid-based path planning discretizes the environment into a fixed grid where each cell represents a specific region in the state space. Sampling-based path planning, on the other hand, avoids explicit discretization of the environment by randomly sampling points within the space. 

Tang et al. \cite{c1} proposed $A^{*}$ algorithm for path planning in port environments and used a logical filter to eliminate irregularities like cross paths and saw-tooth patterns. Bezier curves were applied to smooth turns, but the algorithm's lack of non-holonomic constraint consideration limited its real-world applicability. Esposto et al. \cite{c2} proposed a hybrid dynamic path planning solution for car parking, dividing the problem into three layers: global path planning using $A^{*}$, local path adaptation considering non-holonomic constraints, and trajectory generation. While effective, this approach was primarily tested on parking scenarios. Dolgov et al. \cite{c3} proposed a path-planning algorithm for autonomous vehicles in unknown environments. It applied a kinematics-based variant of $A^{*}$ with a modified state-update rule for kinematic feasibility, followed by numeric non-linear optimization to smooth the path. Liu et al. \cite{c4} discuss algorithms for dynamic obstacle avoidance for autonomous smart cars. Takei et al. \cite{c5} discuss path planning for a car using the Hamilton-Jacobi approach. Sedighi et al. \cite{c6} discuss a path planning algorithm where visibility graphs guide $A^{*}$.

Kavraki et al. \cite{c7} introduced the Probabilistic Roadmap (PRM) method to construct a roadmap by randomly sampling points in the configuration space and connecting them to form a graph of feasible paths. Kuffner et al. \cite{c8} discusses RRT-Connect algorithm to improve efficiency of RRT by growing bi-directional trees. Karaman et al. \cite{c9} introduced RRT*, an asymptotically-optimal incremental sampling-based motion planning algorithm that is guaranteed to converge to an optimal solution. Sucan et al. \cite{c10} introduced KPIECE, a tree-based planner that uses a discretization to guide the exploration of continuous space. Yanbo et al. \cite{c11} introduced Stable Sparse RRT, an asymptotically near-optimal incremental algorithm are shown to converge fast to high-quality paths, while they maintain only a sparse set of samples, which makes them computationally efficient.

A common limitation of instantiations of these methods for mobile robot path planning is reliance on addressing non-holonomicity across multiple layers of path planning. In this paper, we propose path planning for mobile robots that incorporate vehicle dimensions and non-holonomic constraints directly in the $A^{*}$ algorithm layer. Two different implementations of the $A^{*}$ algorithm are proposed, enhanced to account for non-holonomicity and the robot's finite dimensions These two implementations are termed as \textbf{non-holonomic $A^{*}$} and \textbf{geometric $A^{*}$}. Please note the distinction between our geometric $A^{*}$ algorithm and the work of Tang et al. \cite{c1}. Although both algorithms share the name "geometric $A^{}$," they differ significantly in implementation details. Tang et al. refer to their algorithm as geometric $A^{*}$ because they incorporate a filtering layer to address geometric irregularities in the resultant path. In contrast, our algorithm is termed geometric $A^{*}$ because we develop a geometric model that specifically captures the non-holonomic characteristics of the robot.

The remainder of the paper is organized as follows: Section \ref{problem} discusses modelling approaches and problem formulation. Section \ref{path} introduces two path planning algorithms, with their results detailed in Section \ref{sim_results}. Section \ref{conclusion} concludes with discussions on various aspects of these algorithms and outlines future research directions.

\section{PROBLEM FORMULATION}
\label{problem}

The implementation of the $A^{*}$ algorithm involves three essential components:
\begin{enumerate}
  \item \textit{Path Cost Function} 
 that calculates the cost of moving from the current node to its neighboring nodes
  \item \textit{Heuristic distance metric} that determines the anticipated cost from the current node to the goal node
  \item \textit {Neighbour selection} which is necessary for exploration of path
\end{enumerate}

For a point mass robot in a 2D grid environment, the neighbor nodes can be the eight adjacent nodes in the current node's adjacency list. However, for robots with non-holonomic constraints, the set of reachable nodes from current node depends on the vehicle's dimensions, its minimum allowable turn radius, and the current velocity of the vehicle. To account for non-holonomic constraints, the neighbor selection criteria is reformulated, and reachable nodes from the current node are calculated using either the kinematic model of the vehicle or motion primitives that satisfy the minimum turn radius constraint. These two formulations result in two algorithms. 

\subsection{Kinematic Model of Vehicle}
\begin{figure}[h!] 
    \centering
    \includegraphics[angle=0, width=6cm]{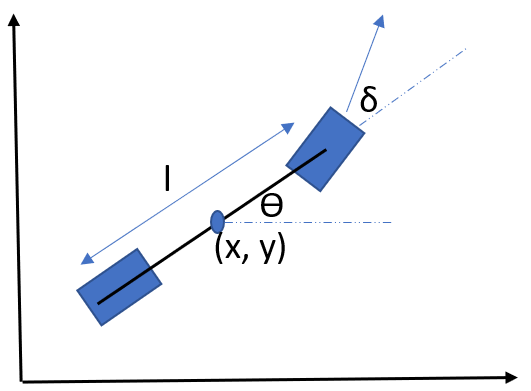}
    \caption{Kinematic model}
    \label{fig:kin_mod}
\end{figure}

A wheeled ground vehicle is a non-holonomic system that has limited steering radius and can only move in a forward or backward direction as shown in figure \ref{fig:kin_mod}. It has no velocity degree of freedom in direction along the wheel axle. Therefore, planning algorithm incorporates kinematic vehicle model to comply with the non-holonomic constraints of the vehicle. The model comprises of position coordinates ($x$, $y$) and heading angle ($\theta$), along with two input variables- velocity ($v$) and steering angle ($\delta$). The configuration of the system is given by tuple of ($x,y,\theta$). 

The kinematic equations are as follows:
\begin{align}
    \dot{x} &= v  \mathrm{cos} \theta \\
    \dot{y} &= v  \mathrm{sin} \theta \\
    \dot{\theta} &= \frac{v}{l} \mathrm{tan}\delta
\end{align}
Here, $l$ denotes the distance between front and rear wheel axles, $v$ denotes the current velocity of the vehicle. In planning algorithm, the length of vehicle is approximated to be $l$.
Euler discretization of the kinematic model is then used to find set of reachable nodes from current state given the current velocity of the robot. This set forms the neighbour nodes for usage in $A^{*}$ algorithm.

\subsection{Geometric Model}

Motion of the vehicle can be equivalently modelled using geometry via abstracting the equation of motion by specifying motion primitives. Mobile Robot cannot move in any random direction, and are limited by their maximum turning angle. In other words, they are limited by their minimum turn radius, and a tighter turn below this is physically not possible. This minimum turn radius defines a circle of motion that the vehicle can follow. 

\begin{figure}[h!] 
    \centering
    \includegraphics[angle=0, width=6cm]{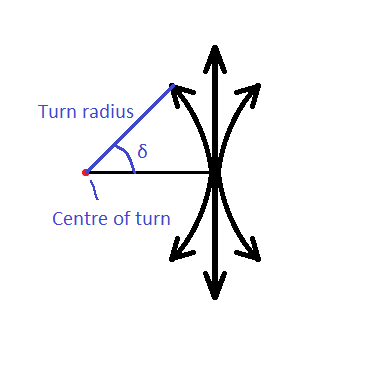}
    \caption{Geometric model}
    \label{fig:geom_model}
\end{figure}

Given some finite turning radius $r$, a centre of turning is obtained by traversing $r$ units in a direction perpendicular to heading angle $\theta$ as shown in figure \ref{fig:geom_model}. Considering polar co-ordinates centred at this centre of turn, the equations of motion is obtained as follows:
\begin{align}
     r_{next} &= r_{current} = r  \label{geom_model1} \\
     \theta_{next} &= \theta_{current} + \delta \label{geom_model2}
\end{align}
where $\delta$ signifies angle of the arc travelled (approximately similar to steering angle, so same notation is used). 
\subsection{Selection of neighbouring nodes}
Neighbour selection is performed using the non-holonomic models presented above. To do so, first velocity, $v$ and current steering angle, $\delta$ is discretized as 
\begin{align*}
    v &= \{[v_1, v_2, v_3, \ldots, v_N] \mid v_i \in [-v_{\text{max}}, v_{\text{max}}]\} \\
    \delta &= \{[\delta_1, \delta_2, \delta_3, \ldots, \delta_N] \mid \delta_i \in [-\delta_{\text{max}}, \delta_{\text{max}}]\}
\end{align*}
where $v_{max},\delta_{max}$ represents maximum robot speed and steering angle respectively. This results in N reachable neighbours from current state, each computed from discretized velocity and steering angle using the equations of motion described above. 
\subsection{Selection of Heuristic Metric}
In the proposed path planning approach, the heuristic cost for a node is not solely based on the euclidean distance to the goal, as conventionally done in search algorithms like $A^{*}$ and Dijkstra. Instead, since the algorithm takes into account the non-holonomic nature of the vehicle, the position of the node is represented by three variables, namely, $(x_{node}, y_{node}, \theta_{node})$, where $\theta$ denotes the heading angle. Therefore, the heuristic function used in this approach incorporates the heading angle in addition to the euclidean distance. The heuristic function can be expressed as:
 \[ h(n) =\sqrt{ (x_n - x_g)^2 + (y_n - y_g)^2 + (\theta_n - \theta_g)^2 } \]
 Here, $\theta$ is expressed in radians.
\subsection{Path Cost Function}
The path cost function used in the proposed algorithm takes into account the actual cost incurred by the robot to reach the current node from the start node. Unlike conventional algorithms, which calculate the sum of distances between the nodes in the path from start node to the current node, the robot in this case takes a curved path if there is a non-zero steering angle. Therefore, an additional cost function ($c(\delta)$) is added to the path cost function to account for the extra distance traveled. To further discourage the robot from making unnecessary turns, this additional cost can be increased as a tune-able parameter.

Next, the direction of the velocity is considered as a parameter to the cost. If the selected node requires a positive velocity (in the vehicle frame of reference), no additional cost ($c(v)$)is considered. However, if the selected node requires a negative velocity, it incurs a penalty in the form of another additional cost. Hence the path cost function is as shown below:
\[ g(n+1) = g(n) + dist.(node_{n+1},node_n) + c(\delta) + c(v) \]
where 'dist' signifies the $L_2$ norm of the difference between the node states $( x, y, \theta)$, similar to the heuristic function.
\section{PATH PLANNING}
\label{path}

An implementation of $A^{*}$ algorithm is provided considering the system model described in Section \ref{problem}. The proposed two variants of $A^{*}$ algorithm built using kinematic and geometric model of vehicle. Implementation can be switched from one variant to another variant by changing the neighbour selection criteria as per both the models. Implementations based on a kinematic model is termed Non-holonomic $A^{*}$ because it explicitly incorporates non-holonomic constraints into the equations of motion. On the other hand, implementation based on a geometric model is termed Geometric $A^{*}$ because it implicitly addresses non-holonomicity by abstracting the motion primitives that satisfy these constraints.

\subsection{Enhanced $A^{*}$ algorithm}
The state of the vehicle is considered in the formulation of algorithm. Hence each node 'n' is defined by $\{x,y,\theta, x_d, y_d, \theta_d\}$ where ($x,y,\theta$) represent the state of vehicle and ($x_d$,$y_d$,$\theta_d$) is the rounded integer value corresponding to ($x$, $y$, $\theta$). Essentially, ($x_d$,$y_d$) represents the midpoint of the discretized grid in 2D space and ($x$,$y$) represents the actual location of the vehicle in the grid. Following data structures (standard in A* implementation) are used in the implementation:
\begin{itemize}
    \item \textit{Open list and visited list}: keeps a dictionary of the following for each node: \{current node (discrete), cost-to-goal, current node (continuous), parent node (discrete), parent node (continuous)\}
    \item  \textit{Open heap}: contains \{cost-to-goal, current node (continuous)\} and is used for sorting the open nodes by cost-to-goal and popping a node as and when it is visited
\end{itemize} 
Algorithm \ref{algorithm} shows the flowchart of the enhanced $A^{*}$ algorithm \cite{c2} with \textit{neighborNodes} updated using strategy presented earlier.
\begin{algorithm}
\label{algorithm}
\caption{$A^{*}$ algorithm}\label{algorithm}
\begin{algorithmic}
\Function{$A^{*}$}{$start, goal, obstacles$}
\State $openset \gets \{ start \}$
\State $gScore(start) \gets 0;$
\State $fScore(start) \!\gets\! gScore(start) + Heuristic(start, goal);$
\While {openset is not empty}
\State $current \gets \textrm{node in openset with lowest fscore;}$
\If {current = goal then}
\State $path \gets ReconstructPath$
\State \textbf{return} path;
\State \textbf{break};
\EndIf
\State remove current from openset;
\For {nbr in neighborNodes(current)}
\If{nbr leads to collision of vehicle with obstacles}
\State \textbf{continue};
\EndIf
\State $gTentative \gets gScore(current) + dist(current, nbr) + c(v) + c(\delta) ;$
\If{$gTentative < gScore(nbr)$}
\State $gScore(nbr) \gets gTentative;$
\State $fScore(nbr) \gets gScore(nbr) + Heuristic(nbr, goal);$
\State add nbr to openset;
\EndIf
\EndFor
\EndWhile
\State \textbf{return} FAILURE
\EndFunction
\end{algorithmic}
\end{algorithm}

\subsection{Collision Detection}
To ensure the vehicle's safety during path planning, it is necessary to take into account the size and shape of the vehicle, as well as the obstacles in the environment. One approach to collision detection would be to check whether the vehicle's center point is in obstacle space. However, this approach does not guarantee collision avoidance.

To address this, vehicle size information is constructed by defining corner points of the convex hull of the robot, representing vertices of the polygonal configuration space of the vehicle. We consider the robot convex hull to be rectangular in this work. At each step of exploration, each of these points are checked for collision. Overall, the incorporation of the vehicle's size and shape into the path-planning algorithm significantly improves its safety and robustness in real-world scenarios.

\section{Simulation Results}
\label{sim_results}
To assess the effectiveness of proposed algorithms, custom 2D navigation scenarios are designed that generate diverse test cases to evaluate their performance. The results presented below demonstrate the incremental improvements achieved through the progressive development of the algorithms.

\subsection{Different Initial Headings}
\begin{figure}
     \centering
     \begin{subfigure}[h]{0.23\textwidth}
         \centering
         \includegraphics[width=5cm, height=4cm]{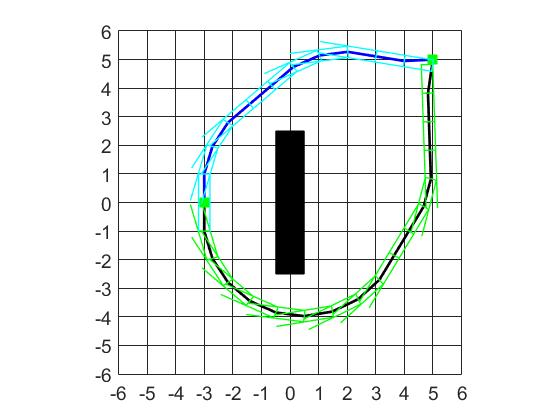}
         \caption{Non-holonomic $A^{*}$}
         \label{Different initial headings 1}
     \end{subfigure}
     \hfill
     \begin{subfigure}[h]{0.23\textwidth}
         \centering
         \includegraphics[width=5cm, height=4cm]{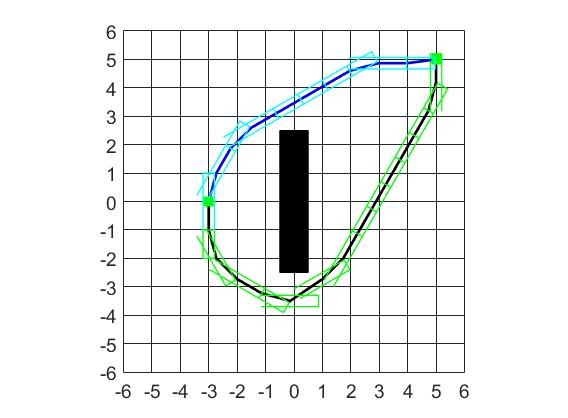}
         \caption{Geometric $A^{*}$}
         \label{Different initial headings 2}
     \end{subfigure}
     \caption{Enhanced A* subject to different initial heading}
     \label{Different initial headings}
\end{figure}

Figure \ref{Different initial headings} illustrates the results of path planning with two different initial vehicle headings, $+90^{\circ}$ (blue curve) and $-90^{\circ}$ (green curve), using Non-holonomic $A^{*}$ algorithm (Fig. \ref{Different initial headings 1}) and Geometric $A^{*}$ algorithm (Fig. \ref{Different initial headings 2}). It is observed that optimal path chosen by both the algorithms vary  taking into account the initial heading of the vehicle. The optimal path obtained in both algorithms have similar trace but slightly different as they incorporate different neighbour selection criteria. 

\subsection{Avoiding Reverse Maneuver}
Optimal solution selection from space of feasible solution often leads to reverse maneuver to re-orient the vehicle towards desired heading. This behavior is particularly beneficial while navigating narrow corridors. Reverse maneuver is directly affected by relative free space available to vehicle and the cost associated with such maneuver. Sometimes, however, reverse maneuvers may be undesirable. One such case may be when a smooth path is required and there is a huge cost associated with the change in direction of the vehicle velocity. When such a maneuver is unwanted, the cost of such reverse direction can be made arbitrarily high to filter out those solutions from space of feasible solutions. This is subject to the availability of free space for navigation relative to the size of vehicle.    

\begin{figure}
     \centering
     \begin{subfigure}[h]{0.23\textwidth}
         \centering
         \includegraphics[width=3.5cm, height=3.75cm]{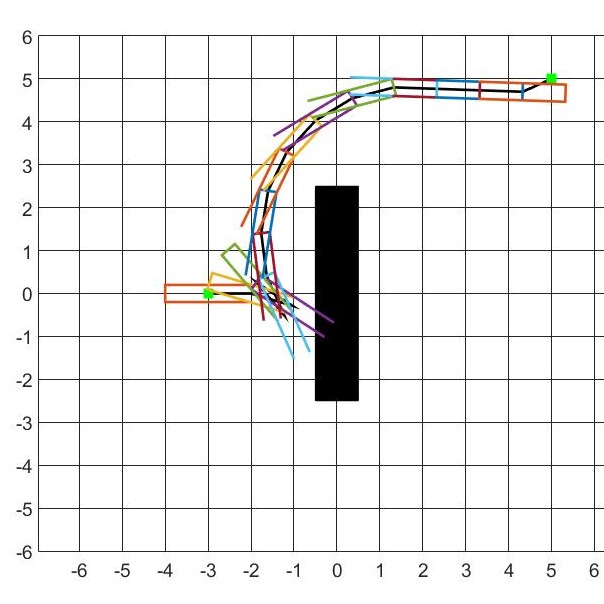}
         \caption{Reverse penalty cost = 1}
         \label{Enabling reverse maneuvering 1}
     \end{subfigure}
     \hfill
     \begin{subfigure}[h]{0.23\textwidth}
         \centering
         \includegraphics[width=4.25cm, height=3.75cm]{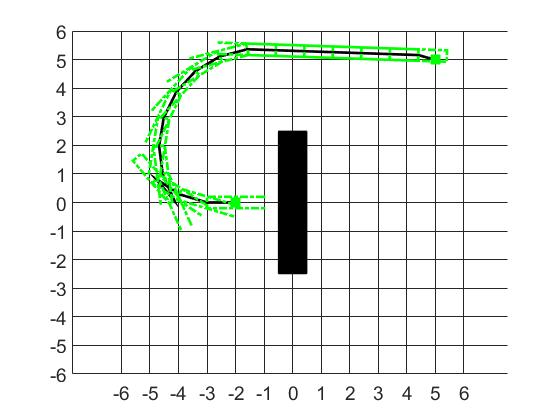}
         \caption{Reverse penalty cost = 100}
         \label{Enabling reverse maneuvering 2}
     \end{subfigure}
     \caption{Controlling reverse manuever by cost selection}
     \label{Enabling reverse manuevering}
\end{figure}

The results of path planning with varying costs assigned to the reverse direction are shown in Fig. \ref{Enabling reverse manuevering}. When the cost of the reverse maneuver is increased, the total number of reverse maneuvers decreases. In some cases, when there is a space limitation, we see that reverse maneuvers are minimized but cannot be completely eliminated.

Overall, the implementation of the reverse maneuvering technique with direction based cost allocation leads to efficient path planning and better maneuverability of the vehicle as per the requirements.

\subsection{U-turn Scenarios}
\begin{figure}
     \centering
     \begin{subfigure}[h]{0.23\textwidth}
         \centering
         \includegraphics[width=5cm, height=3.5cm]{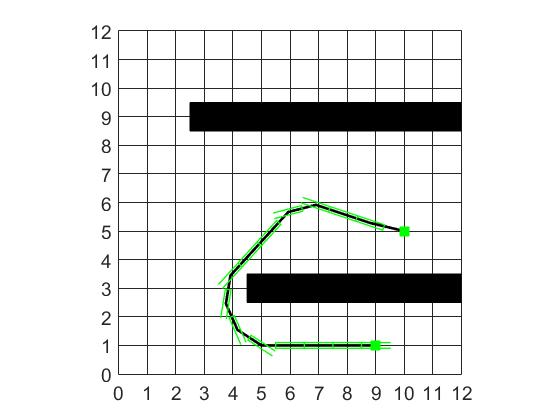}
         \caption{Small vehicle}
         \label{Avoiding unwanted reverse maneuver 1}
     \end{subfigure}
     \hfill
     \begin{subfigure}[h]{0.23\textwidth}
         \centering
         \includegraphics[width=5cm, height=3.5cm]{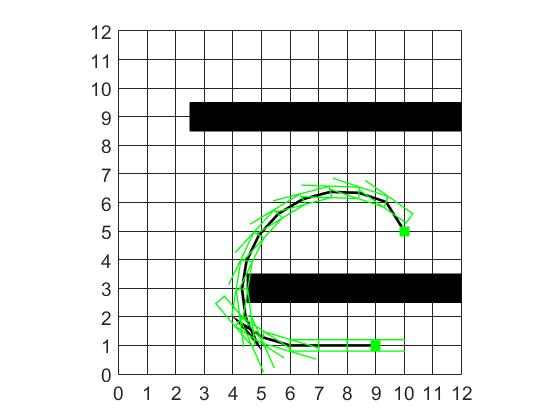}
         \caption{Big vehicle, No penalty}
         \label{Avoiding unwanted reverse maneuver 2}
     \end{subfigure}
     \hfill
     \begin{subfigure}[h]{0.23\textwidth}
         \centering
         \includegraphics[width=5cm, height=3.5cm]{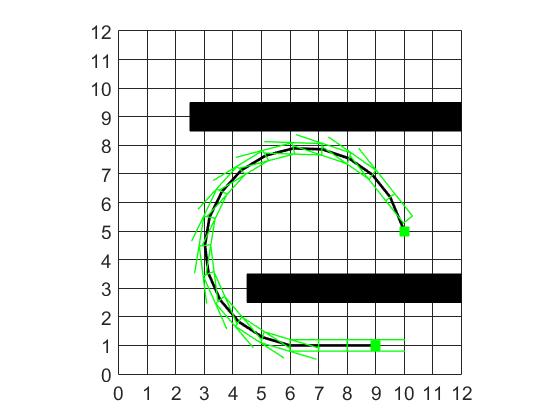}
         \caption{Big vehicle, U-turn penalty}
         \label{Avoiding unwanted reverse maneuver 3}
     \end{subfigure}
     \hfill
     \begin{subfigure}[h]{0.23\textwidth}
         \centering
         \includegraphics[width=5cm, height=3.5cm]{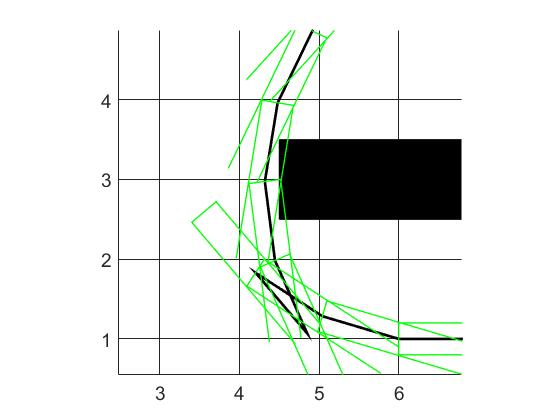}
         \caption{Obstacle collision at U-turn}
         \label{Avoiding unwanted reverse maneuver 4}
     \end{subfigure}
     \caption{Navigation under U-turn scenarios for small (length =1) and big (length = 2) vehicles}
     \label{Avoiding unwanted reverse manuever}
\end{figure}

\begin{figure}[h]
    \centering
    \includegraphics[width=0.5\textwidth]{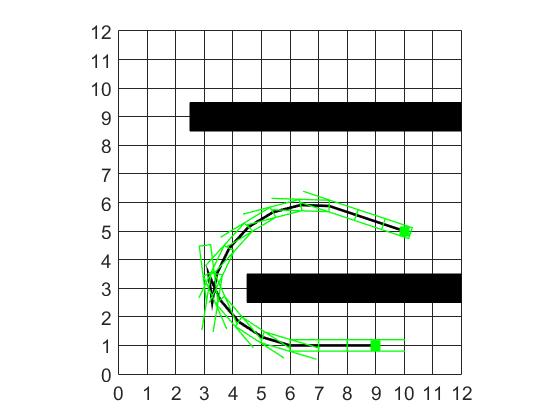}
    \caption{Collision avoidance for the extremities of the vehicle with finite dimensions}
    \label{Collision avoidance finite dimensions}
\end{figure}

Fig. \ref{Avoiding unwanted reverse manuever} demonstrates path planning for a vehicle with different dimensions performing a U-turn. When no cost is incurred for reverse-maneuver, a big vehicle necessarily performs reverse maneuver to re-orient itself as it leads to the least cost path. This is very similar to a truck maneuvering a U-turn in the shortest path. This is avoided by increasing the cost for reverse maneuver as seen in Fig. \ref{Avoiding unwanted reverse manuever}c. 
Fig. \ref{Avoiding unwanted reverse manuever}d illustrates the importance of considering the dimension of vehicle for collision detection. For path planning in Fig. \ref{Avoiding unwanted reverse manuever}d, only the mid-point of the vehicle is checked for collision. This leads to a path where the vehicle outer body collides with obstacles. Fig. \ref{Collision avoidance finite dimensions} demonstrates collision avoidance when the vehicle dimension is taken into account while path planning.

\subsection{Narrow Corridors}
\begin{figure}
     \centering
     \begin{subfigure}[h]{0.23\textwidth}
         \centering
         \includegraphics[width=5.5cm, height=5.5cm]{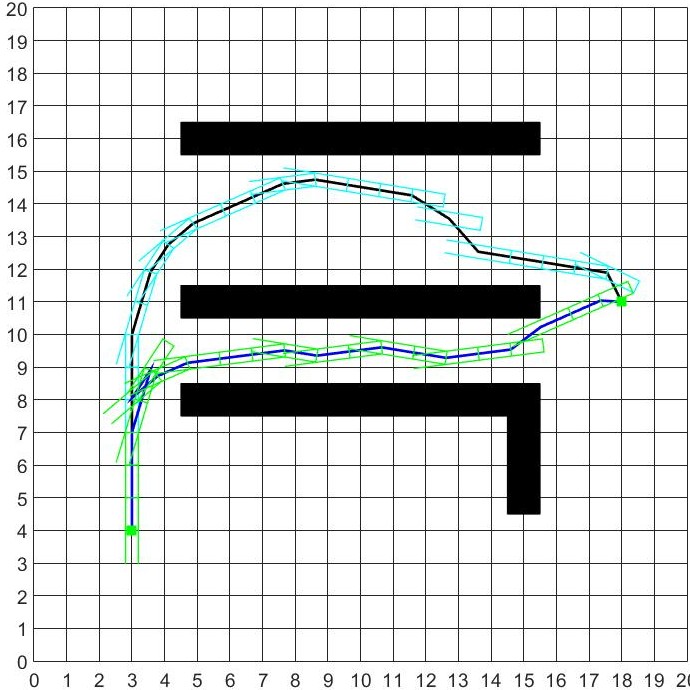}
         \caption{Vehicle with 2 units length}
         \label{Vehicle with length of 2 units}
     \end{subfigure}
     \vfill
     \begin{subfigure}[h]{0.23\textwidth}
         \centering
         \includegraphics[width=5.5cm, height=5.5cm]{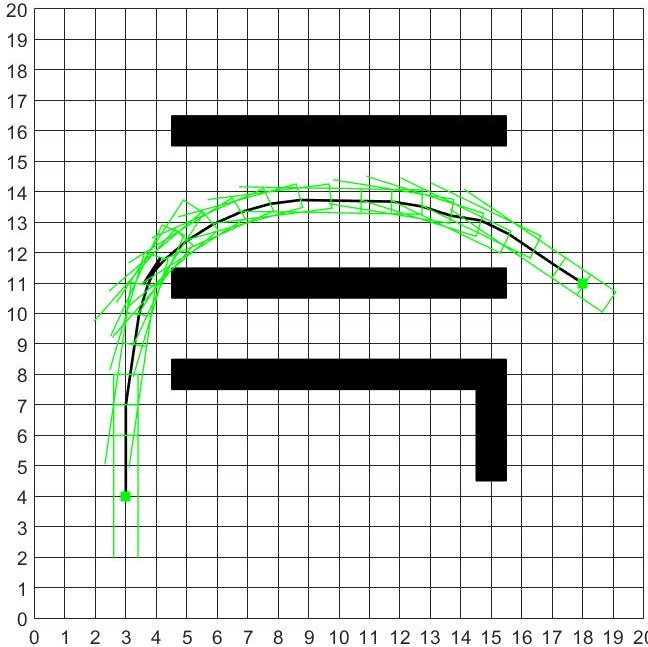}
         \caption{Vehicle with 6 units length}
         \label{Vehicle with length of 6 units}
     \end{subfigure}
     \caption{Test case of narrow lane}
     \label{Real scenario}
\end{figure}

 In Fig. \ref{Real scenario}, simulations are presented that emulate a situation where there may be two possible pathways with different widths leading to the goal state. Fig. \ref{Vehicle with length of 2 units} shows that the vehicle of length 2 units can plan paths from both lanes. For plotting purposes, we represent the vehicle length in multiples of one grid unit for easier visualization. But in order to pass through the narrow lane, it has to make a reverse maneuver to adjust itself and enter the lane without colliding with any obstacles. However, as discussed earlier, this approach results in maneuvering in reverse direction, which may not be feasible in all scenarios. Therefore, a cost associated with reverse direction maneuvering is introduced, which automatically guides the algorithm to choose the wider lane for path planning to avoid reversing. Additionally, in Fig. \ref{Vehicle with length of 6 units}, an illustration of a vehicle with larger dimensions (length = 6 units) is shown, and the algorithm strictly plans a path through the wider lane, as the vehicle cannot enter the narrow lane without leaving the boundary of the 2D space. Incorporating vehicle size in path planning led to better results and prevented the algorithm from selecting paths that may result in collisions.

\begin{figure}
     \centering
     \begin{subfigure}[h]{0.25\textwidth}
         \centering
         \includegraphics[width=6cm, height=5.5cm]{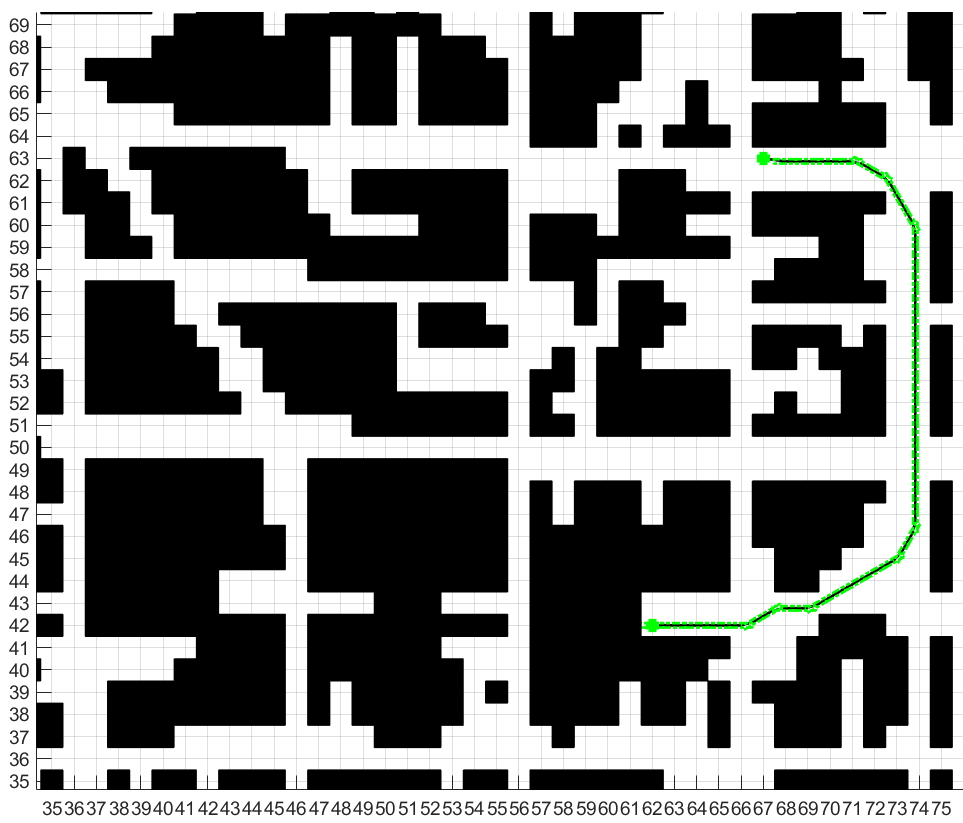}
         \caption{}
         \label{Real scenario 1 1}
     \end{subfigure}
     \vfill
     \begin{subfigure}[h]{0.25\textwidth}
         \centering
         \includegraphics[width=6cm, height=5.5cm]{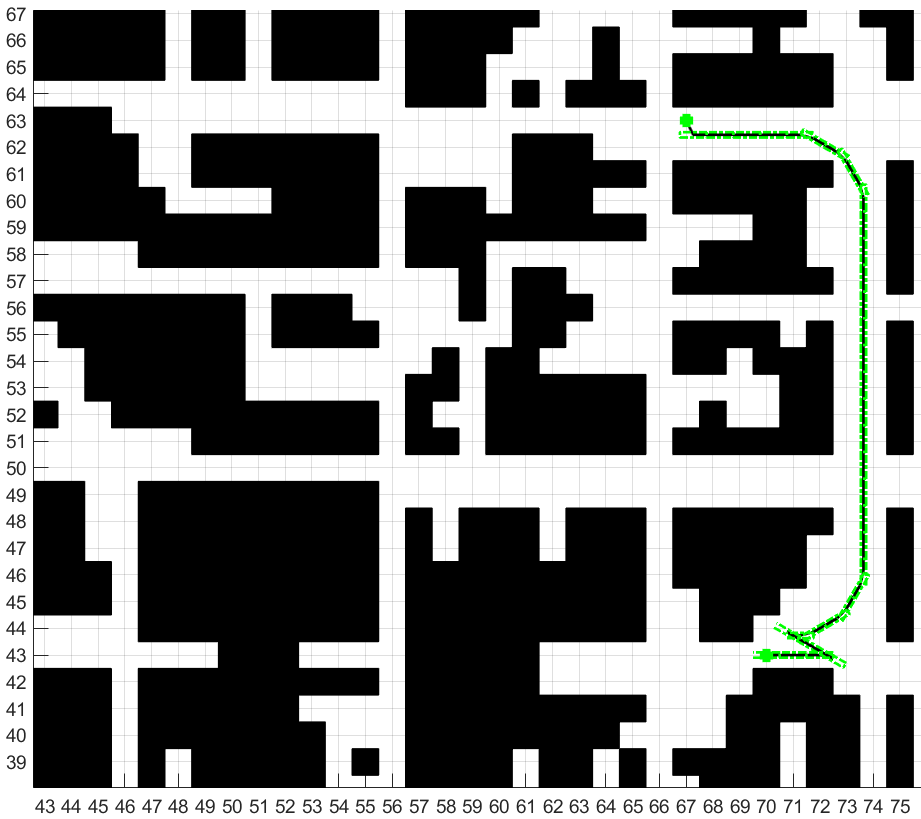}
         \caption{}
         \label{Real scenario 1 2}
     \end{subfigure}
     \caption{Test case of real-world scenario}
     \label{Real scenario 1}
\end{figure}

\subsection{Testing the Algorithm in a Rasterized Map}
To evaluate the performance of the proposed path planning algorithm in a real-world scenario, a section of road network acquired from Google Maps was rasterized to form a binary map. The algorithm was then tested by selecting a start and goal location and obtaining the optimal path between the nodes using the above approach. Implementation of the algorithm on the real-world map is shown in Fig. \ref{Real scenario 1 1}. Here we observe that the optimal path chosen ignores the shortest open path due to vehicle constraints. It is also able to position itself favorably at the bottleneck so that it can easily traverse the remaining path.

Next, the algorithm was tested on a scenario where the vehicle is placed just at the bottleneck along the wall, heading towards East. It cannot immediately make a turn due to its size and dynamic constraints, as depicted in Fig. \ref{Real scenario 1 2}. Despite this challenge, the algorithm was able to plan a path that involves the vehicle adjusting its orientation by making a right turn when possible, followed by a reverse maneuver, effectively avoiding collision with obstacles in its path. These results demonstrate the algorithm's robustness and effectiveness in ensuring the vehicle's safety while navigating through cluttered environments.

\section{CONCLUSIONS}
\label{conclusion}
A novel enhanced A* algorithm has been developed for path planning of mobile robots, incorporating the non-holonomic behavior of vehicles. The proposed algorithm has undergone extensive testing in simulations across various autonomous navigation scenarios. The results demonstrate that the path planning algorithm efficiently integrates non-holonomic constraints and vehicle dimensions to generate optimal paths.

However, comparing the two algorithm instantiations, which consider different models, is not straightforward due to their distinct underlying models for neighbor selection. The turn radius constraint is directly reflected in the kinematic model through factors like axle length and maximum steering angle. In contrast, the geometric model reflects the turn radius constraint through the geometry of circles and their radii. The vehicle's length directly influences the turn radius constraint in the kinematic model, whereas in the geometric model, the formulation is independent of vehicle length and considers length primarily for collision detection. Generally, the kinematic model requires more computation time, while the geometric model is more computationally efficient. We also observed that the kinematic model often produces less smooth paths for U-turns, whereas the geometric model inherently generates smoother paths. 

In the future, we plan to further evaluate these algorithms on more complex real-world maps to assess their robustness, compare the computational complexity against standard other path planning algorithms and extend this approach for path planning of rover for ISRO's next interplanetary mission. Potential improvements include path refinement techniques that consider intermediate steer angles and velocities, incorporating dynamics constraints along with kinematic constraints in planning smooth trajectories for low level controllers. 

\section{Acknowledgments}
We convey our sincere gratitude U.R. Rao Satellite Centre, ISRO for encouraging and supporting this research. We wish to gratefully acknowledge the excellent review committee for reviewing this work and providing valuable feedback. This research work was carried out at Controls and Digital Area in U.R. Rao Satellite Centre.





\end{document}